\def\year{2021}\relax
\documentclass[letterpaper]{article} 
\usepackage{aaai21}  
\usepackage{times}  
\usepackage{helvet} 
\usepackage{courier}  
\usepackage[hyphens]{url}  
\usepackage{graphicx} 
\usepackage{natbib}  
\usepackage{caption} 

\usepackage{adjustbox}
\usepackage{algorithm}
\usepackage{algpseudocode}
\usepackage{calrsfs}
\usepackage[dvipsnames]{xcolor}
\usepackage{multirow}
\usepackage{graphicx}
\usepackage{comment}
\usepackage{amsmath,amssymb} 
\usepackage{color}
\usepackage{subcaption}
\usepackage{verbatim}

\usepackage{xr}
\makeatletter
\newcommand*{\addFileDependency}[1]{
  \typeout{(#1)}
  \@addtofilelist{#1}
  \IfFileExists{#1}{}{\typeout{No file #1.}}
}
\makeatother

\newcommand*{\myexternaldocument}[1]{
    \externaldocument{#1}
    \addFileDependency{#1.tex}
    \addFileDependency{#1.aux}
}
\myexternaldocument{supp_ver3}

\title{\text{LADA}: Look-Ahead Data Acquisition via Augmentation for Active Learning}
\author{Yoon-Yeong Kim, Kyungwoo Song, JoonHo Jang, Il-Chul Moon\\}
\affiliations{Department of Industrial and Systems Engineering \\
	KAIST\\
	Daejeon, Republic of Korea \\
	\texttt{\{yoonyeong.kim,gtshs2,adkto8093,icmoon\}@kaist.ac.kr} \\
}

\begin{document}
\maketitle

\begin{abstract}
\textit{Active learning} effectively collects data instances for training deep learning models when the labeled dataset is limited and the annotation cost is high.
Besides active learning, \textit{data augmentation} is also an effective technique to enlarge the limited amount of labeled instances. However, the potential gain from virtual instances generated by data augmentation has not been considered in the acquisition process of active learning yet.
Looking ahead the effect of data augmentation in the process of acquisition would select and generate the data instances that are informative for training the model.
Hence, this paper proposes Look-Ahead Data Acquisition via augmentation, or \text{LADA}, to integrate data acquisition and data augmentation. \text{LADA} considers both 1) unlabeled data instance to be selected and 2) virtual data instance to be generated by data augmentation, in advance of the acquisition process.
Moreover, to enhance the informativeness of the virtual data instances, \text{LADA} optimizes the data augmentation policy to maximize the predictive acquisition score, resulting in the proposal of \textit{InfoMixup} and \textit{InfoSTN}.
As \text{LADA} is a generalizable framework, we experiment with the various combinations of acquisition and augmentation methods.
The performance of \text{LADA} shows a significant improvement over the recent augmentation and acquisition baselines which were independently applied to the benchmark datasets.
\end{abstract}

\section{Introduction}
Large-scale datasets in the big data era have opened the blooming of artificial intelligence, but the data labeling requires significant efforts from human annotators. Therefore, an adaptive sampling, i.e. \textit{Active Learning}, has been developed to select the most informative data instances in learning the decision boundary \cite{cohn1996active,tong2001active,settles2009active}.
This selection is difficult because it is influenced by the learner and the dataset at the same time. Hence, the understanding of the relation between the learner and the dataset has become the components of \textit{active learning}, which queries the next training example by the informativeness for learning the decision boundary.

\begin{figure}[t]
\captionsetup[subfigure]{justification=centering}
\centering
\begin{subfigure}{.49\linewidth}
    \centering 
    \includegraphics[width=40mm,height=21mm]{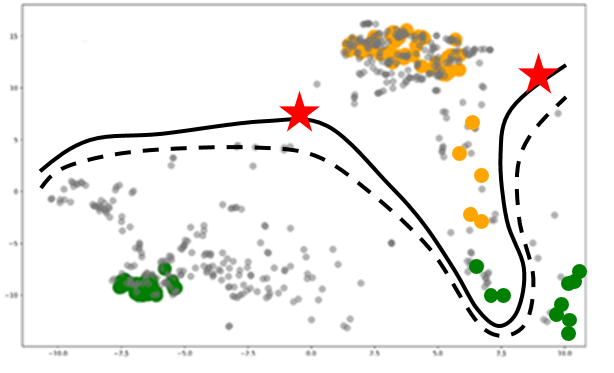}
    \vspace{-1mm}
    \caption{\textit{Max Entropy} \\ \quad}
    \label{fig_IntroAcq}
\end{subfigure}
\begin{subfigure}{.49\linewidth}
    \centering 
    \includegraphics[width=40mm,height=21mm]{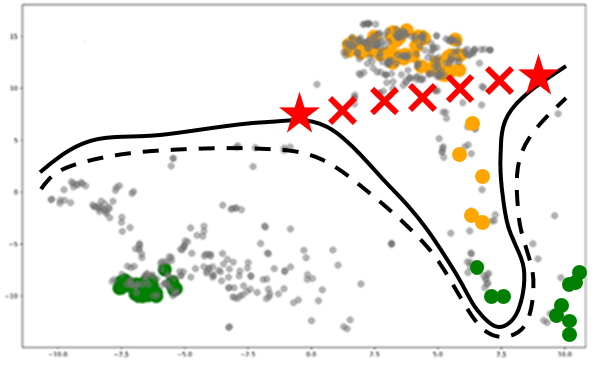}
    \vspace{-1mm}
    \caption{\textit{Mixup} applied \\ after \textit{Max Entropy}}
    \label{fig_IntroAcqDA}
\end{subfigure}

\begin{subfigure}{.49\linewidth}
  \centering
  \includegraphics[width=40mm,height=21mm]{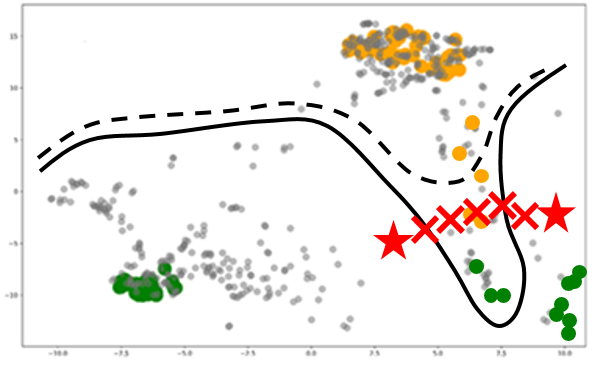}
    \vspace{-1mm}
  \caption{\text{LADA} without \\ learnable augmentation}
  \label{fig_IntroLAfix}
\end{subfigure}
\begin{subfigure}{.49\linewidth}
  \centering
  \includegraphics[width=40mm,height=21mm]{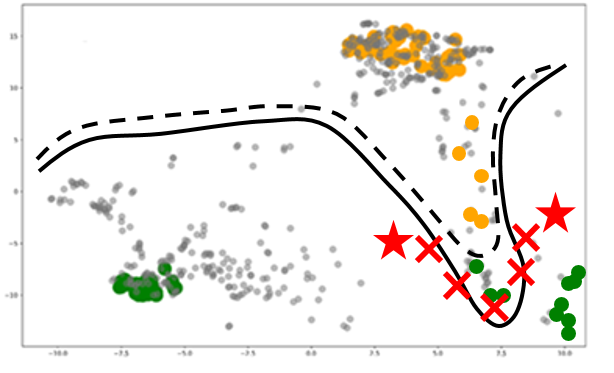}
    \vspace{-1mm}
  \caption{\text{LADA} with \\ learnable augmentation}
  \label{fig_IntroLA}
\end{subfigure}
\vspace{-2mm}
\caption{Illustration of different scenarios for applying acquisition and augmentation; with selected instances ($\star$), virtual instances ($\times$), current decision boundary (solid line), and updated decision boundary (dashed line). (a) \textit{Max Entropy} selects the instances near the decision boundary and updates the boundary. (b) If we augment the selected instances afterward, the virtual instances may not be useful for updating the boundary. (c) By considering the potential gain of the virtual instances from augmentation in advance of the acquisition, we enhance the informativeness of both the selected and augmented instances for learning the decision boundary. (d) Moreover, we train the augmentation policy to maximize the acquisition score of the virtual instances to generate useful instances.}
\label{fig_Intro}
\end{figure}

Besides \textit{active learning}, \textit{data augmentation} is another source of providing virtual data instances to train models. The labeled data may not cover the full variation of the generalized data instances, so the learner has used the \textit{data augmentation}; particularly in the vision community \cite{poseestimation, perez2017effectiveness, autoaugment}.
Conventional data augmentation has been a simple transformation of labeled data instances, i.e. flipping, rotating, \textit{etc}. Recently, the data augmentation has expanded to become a deep generative model, such as Generative Adversarial Networks (GAN) \cite{GAN} or Variational Autoencoder (VAE) \cite{VAE}, that generate virtual examples. Since the conventional augmentations and the generative model-based augmentations perform the Vicinal Risk Minimization (VRM)\cite{VRM}, they assume that the virtual data instances in the vicinity share the same label, which leads to limiting the feasible vicinity. To overcome the limited vicinity of VRM, \textit{Mixup} and its variants have been proposed by interpolating multiple data instances \cite{Mixup}. The pair of interpolated features and labels, or the \textit{Mixup} instance, becomes a virtual instance to enlarge the support of the training distribution.

\textit{Data augmentation} and \textit{active learning} intend to overcome the scarcity of labeled dataset in different directions.
First, \textit{active learning} emphasizes the optimized selection of the unlabeled real-world instances for the \textit{oracle} query, so there is no consideration on the benefit of the virtual data generation. Second, the \textit{data augmentation} focuses on generating an informative virtual data instance without intervening on the data selection stage, and without the potential assistance from \textit{oracle}.
These differences motivate us to propose the Look-Ahead Data Acquisition via augmentation, or \text{LADA} framework.

\text{LADA} looks ahead the effect of data augmentation in advance of the acquisition process, and \text{LADA} selects data instances by considering both unlabeled data instances and virtual data instances generated by data augmentation, at the same time. Whereas the conventional acquisition function does not consider the potential gain of the data augmentation, \text{LADA} contemplates the informativeness of the virtual data instances by integrating data augmentation into the acquisition process. Figure \ref{fig_Intro} describes the different behavior of \text{LADA} and conventional acquisition functions when applying data augmentation to active learning.

Here are our contributions from the methodological and the experimental perspectives.
First, we propose a generalized framework, named \text{LADA}, that looks ahead the acquisition score of the virtual data instance to be augmented, in advance of the acquisition. 
Second, we train the data augmentation policy to maximize the acquisition score to generate informative virtual instances. Particularly, we propose two data augmentation methods, \textit{InfoMixup} and \textit{InfoSTN}, which are trained by the feedback of acquisition scores.
Third, we substantiate the proposed framework by implementing the variations of acquisition-augmentation frameworks with known acquisitions and augmentation methods.
\vspace{-0.7mm}

\section{Preliminaries}
\vspace{-0.3mm}
\subsection{Problem Formulation}
This paper trains a classifier network, $f_{\theta}$, with dataset $\mathcal{X}$ while our scenario is differentiated by assuming $\mathcal{X}=\mathcal{X}_U \cup \mathcal{X}_L$ and $|\mathcal{X}_U| \gg |\mathcal{X}_L|$. Here, $\mathcal{X}_U$ is a set of unlabeled data instances, and $\mathcal{X}_L$ is a labeled dataset.
Given these notations, a data augmentation function, $f_{aug}(x;\tau) \colon \mathcal{X} \to V(\mathcal{X})$, transforms a data, $x \in \mathcal{X}$, into a modified data, $\tilde{x} \in V(\mathcal{X})$; where $\tau$ is a parameter describing the policy of transformation, and $V(\mathcal{X})$ is the vicinity set of $\mathcal{X}$ \cite{VRM}. On the other hand, a data acquisition function, $f_{acq}(x;f_{\theta}) \colon \mathcal{X}_{U} \to \mathbb{R}$, calculates a score of each data instance, $x \in \mathcal{X}_{U}$, based on the current classifier, $f_{\theta}$; and $f_{acq}$ represents the instance selection strategy in the learning procedure of $f_{\theta}$ with the instance, $x \in \mathcal{X}_U$.

\subsection{Data Augmentation}
In the conventional data augmentations, $\tau$ in $f_{aug}(x;\tau)$ indicates the predefined degree of rotating, flipping, cropping, etc. $\tau$ is manually chosen by the modeler to describe the vicinity of each data instance.

Another approach of modeling $\tau$ is utilizing the feedback from the current classifier network, $f_{\theta}$. Spatial Transformer Network (STN) is a transformer to generate a virtual example by training $\tau$ to minimize the cross-entropy (CE) loss of the transformed data \cite{STN}:
\begin{equation}
\tau^{*} = \operatorname*{argmin}_{\tau} CE(f_{\theta}(f^{STN}_{aug}(x;\tau)), y),
\end{equation}
where $y$ is the ground-truth label of the data instance, $x$.

Recently, \textit{Mixup}-based data augmentations generate a virtual data instance from the vicinity of a pair of data instances. In \textit{Mixup}, $\tau$ becomes the mixing policy of two data instances, $x_i$ and $x_j$ \cite{Mixup}:
\begin{equation}
\label{eq:mixup}
f^{Mixup}_{aug}(x_i,x_j;\tau) = \lambda x_i + (1-\lambda) x_j, \lambda\sim \text{Beta}(\tau,\tau),
\end{equation}
where the labels are also mixed by the proportion $\lambda$.
While Eq.\eqref{eq:mixup} corresponds to the input feature mixture, \textit{Manifold Mixup} mixes the hidden feature maps from the middle of neural networks to learn smoother decision boundary at multiple levels of representations \cite{ManifoldMixup}.
Whereas $\tau$ is a fixed value without any learning process, \textit{AdaMixup} learns $\tau$ by adopting a discriminator, $\varphi^{ada}$ \cite{AdaMixup}:
\begin{align}
\tau^{*} &= \operatorname*{argmax}_{\tau}\log \text{P}(\varphi^{ada}(f^{Mixup}_{aug}(x_i,x_j;\tau))=1) \nonumber \\
&+ \log \text{P}(\varphi^{ada}(x_i)=0)+\log \text{P}(\varphi^{ada}(x_j)=0).
\end{align}

\subsection{Active Learning}
We focus on the pool-based active learning with uncertainty score \cite{settles2009active}. Given this scope of active learning, the data acquisition function measures the utility score of the unlabeled data instances, i.e. $x^{*} = \operatorname*{argmax}_{x} f_{acq}(x;f_{\theta}).$

The traditional acquisition functions measure the predictive entropy, $f_{acq}^{Ent}(x;f_{\theta}) = \mathbb{H}[y|x;f_\theta]$ \cite{maxentropy}; or the variation ratio, $f_{acq}^{Var}(x;f_{\theta}) = 1 - max_{y}\text{P}(y|x;f_{\theta})$ \cite{VarRatio}. The recent acquisition function calculates the hypothetical disagreement by $f_{\theta}$ on a data instance, $f_{acq}^{BALD}(x;f_{\theta}) = \mathbb{H}[y|x;f_{\theta}]-\mathbb{E}_{\text{P}(\theta|D_{train})}[\mathbb{H}[y|x;f_{\theta}]]$ \cite{BALD}.

Besides the classifier network, $f_{\theta}$, additional modules are applied to measure the acquisition score. To find the most dissimilar instance in $\mathcal{X}_{U}$ compared to $\mathcal{X}_L$, a discriminator, $\varphi^{VAAL}$, is introduced to estimate the probability of belonging to $\mathcal{X}_{U}$ \cite{vaal}:
\begin{align}
f_{acq}^{VAAL}(x;\varphi^{VAAL}) = \text{P}(x\in\mathcal{X}_U;\varphi^{VAAL}).
\end{align}
To diversely select uncertain data instances, the gradient embedding from the pseudo label, $\hat{y}$, is used in \textit{k}-MEANS++ seeding algorithm \cite{badge}:
\begin{align}
f_{acq}^{BADGE}(x;f_\theta)=\frac{\partial}{\partial \theta_{out}}CE(f_{\theta}(x),\hat{y}).
\end{align}

\subsection{Active Learning with Data Augmentation}
There are a few prior works in integrating active learning and data augmentation effectively. Bayesian Generative Active Deep Learning (BGADL) integrates acquisition and augmentation by selecting data instances via $f_{acq}$, then augmenting the selected instances via $f_{aug}$, which is VAE, afterward \cite{BGADL}. However, BGADL limits the vicinity to preserve the labels, and BGADL demands on large labeled instances to train generative models. More importantly, BGADL does not consider the potential gain of data augmentation in the process of acquisition.

\begin{figure*}[t]\label{fig_LADA}
\captionsetup[subfigure]{justification=centering}
  \centering
  \begin{subfigure}{.49\linewidth}
  \includegraphics[width=\linewidth]{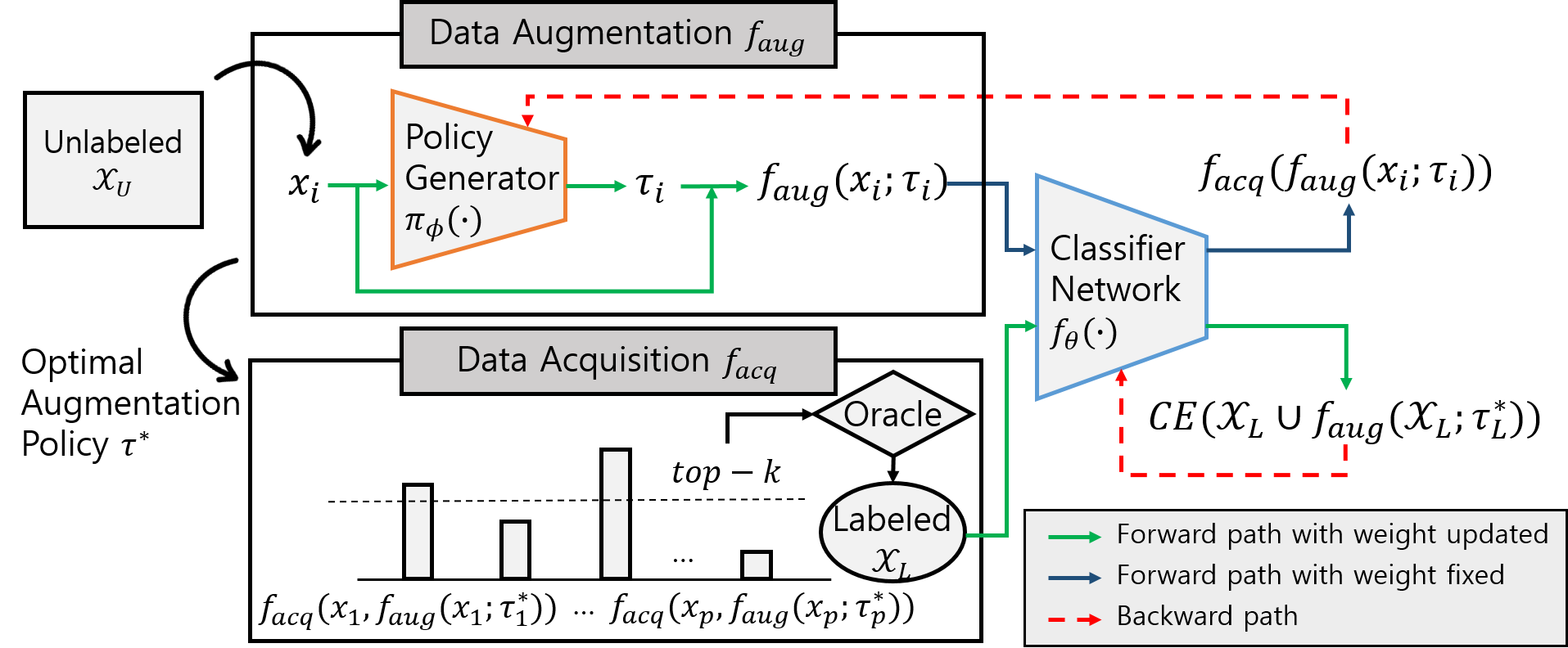}
  \caption{Overview of \text{LADA} framework}
  \label{fig_Overview}
  \end{subfigure}
  \begin{subfigure}{.49\linewidth}
  \centering
  \includegraphics[width=\linewidth]{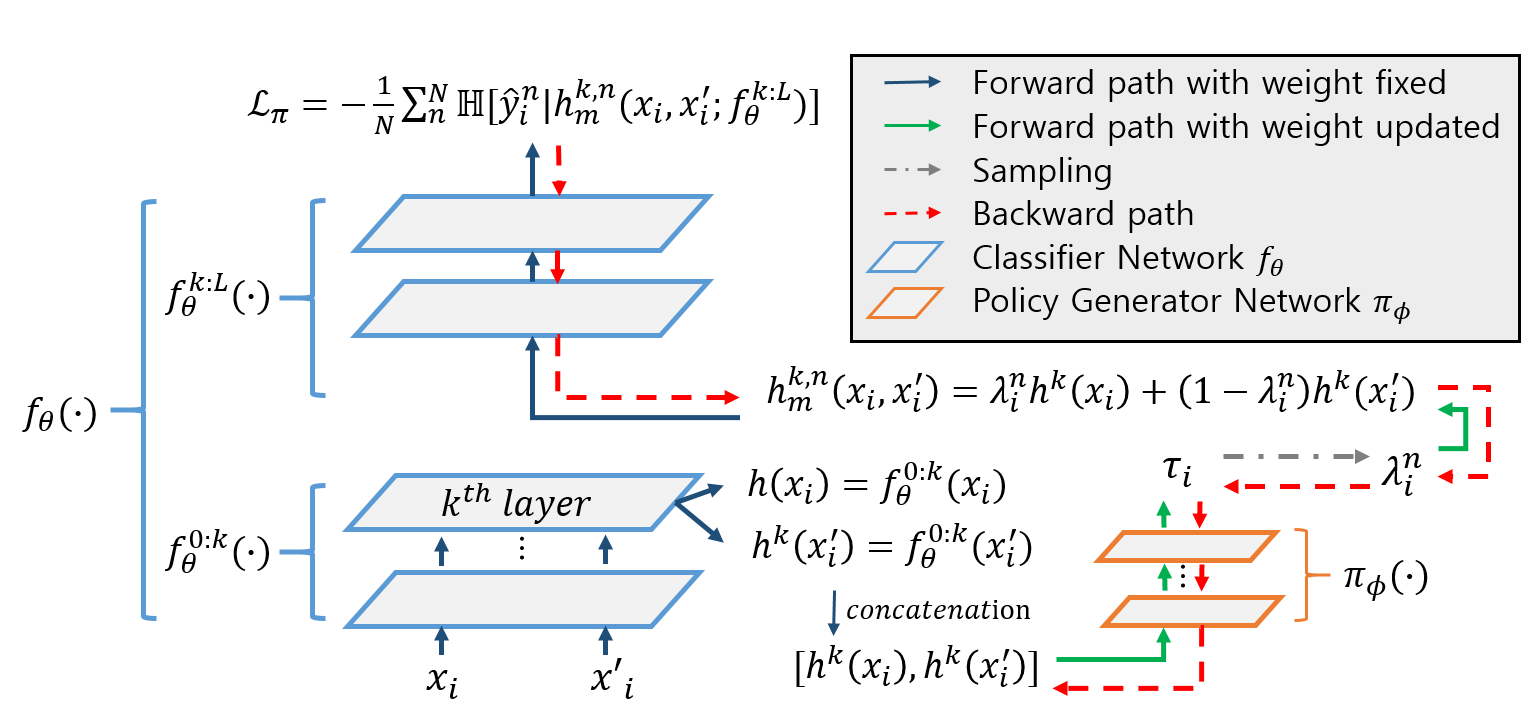}
  \caption{Training process of the policy generator network, $\pi_{\phi}$, \\ in \text{LADA} with \textit{Max Entropy} and \textit{Manifold Mixup}}
  \label{fig_PGN}
  \end{subfigure}
  \caption{Overall algorithm of \text{LADA} and training process of data augmentation. (a) $f_{acq}$ considers both the unlabeled instances and the virtual instances generated from $f_{aug}$. Moreover, $f_{aug}$ is optimized with the feedback of $f_{acq}$. (b) The parameters of the classifier network $f_{\theta}$ are fixed during the backpropagation for the policy generator network.}
\end{figure*}

\section{Methodology}
A contribution of this paper is proposing an integrated framework of data augmentation and acquisition, so we start from formulating such a framework. Afterward, we propose an integrated function for acquisition and augmentation as an example of the implemented framework.

\subsection{Look-Ahead Data Acquisition via Augmentation}
Since we look ahead the acquisition score of augmented data instances, it is natural to integrate the functionalities of acquisitions and augmentations. This paper proposes a Look-Ahead Data Acquisition via augmentation, or \text{LADA} framework.
Figure \ref{fig_Overview} depicts the \text{LADA} framework, which consists of the data augmentation component and the acquisition component.
The goal of \text{LADA} is enhancing the informativeness of both 1) real-world data instance, which is unlabeled at current, but will be labeled by the \textit{oracle} in the future; and 2) virtual data instance, which will be generated from the unlabeled data instances that are selected. This goal is achieved by looking ahead of their acquisition scores before actual selections for the \textit{oracle} annotations.

Specifically, \text{LADA} trains the data augmentation function, $f_{aug}(x;\tau)$, to maximize the acquisition score of the transformed data instance of $x_U$ before the \textit{oracle} annotations. Eq.\eqref{eq:aug_acq} specifies the learning objectives of the augmentation policy via the feedback from acquisition.
\begin{equation}
\label{eq:aug_acq}
\tau^{*} = \operatorname*{argmax}_{\tau} f_{acq}(f_{aug}(x_{U};\tau);f_{\theta}).
\end{equation}
With the optimal $\tau^{*}$ corresponding to $x_U$, $f_{acq}$ calculates the acquisition score of $x_U$ (see Eq.\eqref{eq:acq_lada}), and the score also considers the utility of the augmented instance, $f_{aug}(x_{U};\tau^{*})$:
\begin{equation}
\label{eq:acq_lada}
x_{U}^{*} = \operatorname*{argmax}_{x_U\in\mathcal{X}_U} [f_{acq}(x_{U};f_{\theta})+f_{acq}(f_{aug}(x_{U};\tau^{*});f_{\theta})]
\end{equation}

Whereas the proposed \text{LADA} framework is a generalized framework that can adopt the various types of acquisition and augmentation functions, this section mainly adopts \textit{Mixup} for $f_{aug}$, i.e. $f_{aug}^{Mixup}$ and \textit{Max Entropy} for $f_{acq}$, i.e. $f_{acq}^{Ent}$. To begin with, we introduce an integrated single function to substitute the composition of functions as $f_{integ}=f_{acq} \circ f_{aug}(x_{U})=f_{acq}(f_{aug}(x_{U};\tau);f_{\theta})$ for generality.

\subsection{Integrated Augmentation and Acquisition: \textit{InfoMixup}}
As we introduce \text{LADA} with $f_{integ}$ to look ahead the acquisition score of the virtual data instances, $f_{integ}$ can be a simple composition of well-known acquisition functions and augmentation functions where the policy of augmentation is fixed. However, this does not enhance the informativeness of the virtual data instances. Hence, we propose the integration where the policy of data augmentation is optimized to maximize the acquisition score, within a single function. Here, we introduce \textit{InfoMixup} as a learnable data augmentation.

\subsubsection{Data Augmentation}
First, we propose \textit{InfoMixup}, which is an adaptive version of \textit{Mixup} to integrate the data augmentation into active learning.
\textit{InfoMixup} learns its mixing policy, $\tau$, by the objective functions Eq.\eqref{eq:infomixup}, where $\lambda \sim \text{Beta}(\tau^{*},\tau^{*})$ maximizes the acquisition score of the virtual data instance resulting from mixing two randomly paired data instances, $x_{i}$ and $x'_{i}$:
\begin{align}
&f_{EntMix}(x_i,x'_i;\tau,f_{\theta}) = f_{acq}^{Ent}(f_{aug}^{Mixup}(x_{i},x'_{i};\tau);f_{\theta})\nonumber\\
&\tau^{*} = \operatorname*{argmax}_{\tau} f_{EntMix}(x_i,x'_i;\tau_,f_{\theta}).\label{eq:infomixup}
\end{align}
\textit{InfoMixup} is the starting ground where we correlate the data augmentation guided by the data acquisition from the perspective of the predictive classifier entropy. 

We adopt \textit{Manifold Mixup} as the data augmentation at the hidden layer. Specifically, the pair of $(x_i,x'_i)\in \mathcal{X}_U$ is processed through the current classifier network, $f_{\theta}$, until the propagation reaches the randomly selected $k$-th layer\footnote{Throughout this paper, we denote the forward path from the $i^{th}$ layer to the $j^{th}$ layer of the classifier network as $f_{\theta}^{i:j}$, where $0$-th is the input layer and $L$-th is the output layer. Hence, $f_{\theta}=f_{\theta}^{0:L}$.}. Afterwards, the $k$-th feature maps $(h^{k}(x_{i}), h^{k}(x'_{i}))$ are concatenated and processed by the policy generator network, $\pi_{\phi}$, to predict $\tau_{i}^{*}$ that maximizes the acquisition score.

\subsubsection{Data Augmentation Policy Learning}
As we formulate the \textit{Mixup} based augmentation, we propose a policy generator network, $\pi_{\phi}$, to perform the amortized inference on the Beta distribution of \textit{InfoMixup}. While we provide the details of the policy network in Appendix A.2 and Figure \ref{fig_PGN}, we formulate this inference process as Eq.\eqref{q;tilde_x} and Eq.\eqref{q;alpha2}.
\begin{align}
h^{k}(x_{i}) &= f_{\theta}^{0:k}(x_{i}), h^{k}(x'_i) = f_{\theta}^{0:k}(x'_{i}) \label{q;tilde_x}\\
\tau_{i} &= \pi_{\phi}([h^{k}(x_{i}),h^{k}(x'_i)])\label{q;alpha2}\\
&=MLP_{\phi}([h^{k}(x_{i}),h^{k}(x'_i)]). \nonumber
\end{align}
To train the parameters, $\phi$, of the policy generator network, $\pi$, the paired features are mixed-up with $N$ sampled $\lambda_{i}$'s. Using a $n$-{th} sampled $\lambda_{i}^{n}$ from the Beta distribution inferred by $\pi_{\phi}$, the feature maps $h^{k}(x_{i})$ and $h^{k}(x'_{i})$ are mixed to produce $h^{k,n}_{m}(x_{i},x'_{i})$ as the below:
\begin{align}
\lambda_{i}^{n} &\sim \text{Beta}(\tau_{i},\tau_{i}), \label{q;lambda}\\
h^{k,n}_{m}(x_i,x'_i) &= \lambda_{i}^{n}h^{k}(x_i)+(1-\lambda_{i}^{n})h^{k}(x'_{i}). \label{q;g}
\end{align}
By processing $h^{k,n}_{m}$ for the rest layers of the classifier network, the predictive class probability of the mixed features is obtained as $\hat{y}_{i}^{n}=f_{\theta}^{k:L}(h^{k,n}_{m}(x_i,x'_{i}))$.
In order to generate a useful virtual instance through \textit{InfoMixup}, the policy generator network has a loss function to minimize the negative value of the predictive entropy as Eq.\eqref{q;L_pi}, and the predictive entropy is a component of $f_{acq}$, which provides the incentive for the integration of acquisition and augmentation. The gradient of this loss function is calculated by averaging the $N$ entropy values of the replicated mixed features.
It should be noted that Eq.\eqref{q;L_pi} embed $\phi$ in the generation of $h^{k,n}_{m}(x_i,x'_{i})$, so the gradient can be estimated via the Monte-Carlo sampling \cite{hastings1970monte}. Figure \ref{fig_PGN} illustrates the forward and the backward paths for the training process of the policy generator network.
\begin{align}
\frac{\partial}{\partial \phi} L_{\pi} = \frac{\partial}{\partial \phi} &(-{1 \over N}\sum_{n}^{N}f_{acq}(h^{k,n}_{m}(x_i,x'_i);f_{\theta}^{k:L})) \label{q;L_pi}\\
 = \frac{\partial}{\partial \phi} &(-{1 \over N}\sum_{n}^{N}\mathbb{H}[\hat{y}_{i}^{n}|h^{k,n}_{m}(x_i,x'_i);f_{\theta}^{k:L}])\nonumber
\end{align}
In the backpropagation, we have a process of sampling $\lambda_{i}$s from the \text{Beta} distribution parameterized by $\tau_{i}$. To enable the backpropagation signals to pass by, we follow the reparameterization technique of the optimal mass transport (OMT) gradient estimator, which utilizes the implicit differentiation \cite{jankowiak2018pathwise, JankowiakK19}. Appendix B provides the details of our OMT gradient estimator in the backpropagation process.

\subsubsection{Data Acquisition by Learned Policy}
After optimizing the mixing policy, $\tau_{i}^{*}$, for $i$-{th} pair of unlabeled data instances, $(x_{i},x'_{i})$, we calculate the joint acquisition score of the data pair by aggregating the individual acquisition scores of 1) $x_{i}$, 2) $x'_{i}$, and 3) their mixed feature maps, $h_{m}^{k,n}(x_i,x'_i)$ as the below:
\begin{align}
\hat{y}_i=f_{\theta}(x_i), \hat{y}'_{i}=f_{\theta}&(x'_{i}), \hat{y}_{i}^{n}=f_{\theta}^{k:L}(h_{m}^{k,n}(x_i,x'_i)) \label{q;y_hat}\\
f_{acq}\big((x_i,x'_i);f_{\theta}\big)&=\mathbb{H}[\hat{y}_{i}|x_i;f_{\theta}]+\mathbb{H}[\hat{y'}_{i}|x'_i;f_{\theta}] \label{q;score}\\
&+{1 \over N}\sum_{n}^{N}\mathbb{H}[\hat{y}_{i}^{n}|h^{k,n}_{m}(x_{i},x'_{i});f_{\theta}^{k:L}]\nonumber
\end{align}
As we calculate the acquisition score by including the predictive entropy of the \textit{InfoMixup} feature map, the acquisition is influenced by the data augmentation. More importantly, this integration is reciprocal because the optimal augmentation policy of \textit{InfoMixup} comes from the acquisition score. This reciprocal relation is an example of motivating the \text{LADA} framework by overcoming the separation between the augmentation and the acquisition.

If we take \textit{InfoMixup} as an example of \text{LADA}, we show that \textit{InfoMixup} generates a virtual sample with a high predictive entropy in the class estimation, which could be regarded as a decision boundary region that is not clearly explored, yet. The unexplored region is identified by the optimal policy of $\tau^*$ in the acquisition. 

Here, we introduce a pipelined variant in $f_{integ}$ to emphasize the worth of the integration. 
One possible variation is incorporating $\textit{Mixup}$-based data augmentation and acquisition function as a two-step model, where 1) the acquisition function selects the data instances whose individual scores are the highest, 2) then \textit{Mixup} is afterward applied to the selected instances. However, this method may increase the criteria of individual data instances laid in the first and the second terms of Eq.\eqref{q;score}, but it may not optimize the criteria of their mixing process laid in the last part of Eq.\eqref{q;score} since it has not considered the effect of $\textit{Mixup}$ in the selection process. This may not enhance the informativeness of the virtual data instances. We compare this variation with \text{LADA} in the Experiments Section.

\newcommand{\factorial}{\ensuremath{\mbox{\sc Factorial}}}
\begin{algorithm}[t]
\caption{\text{LADA} with \textit{Max Entropy} and \textit{Manifold Mixup}}\label{algorithm_InfoMixupAcq}
\renewcommand{\algorithmicrequire}{\textbf{Input:}}
\begin{algorithmic}[1]
\Require{Labeled dataset $\mathcal{X}_{L}^{0}$, Classifier $f_{\theta}$}
\For {$j = 0,1,2,\ldots$} \textcolor{gray}{\Comment{active learning}}
\State Randomly sample $\mathcal{X}_{U}^{pool} \subset \mathcal{X}_{U}$
\State Get $\mathcal{X'}_{U}^{pool}$ which is randomly shuffled $\mathcal{X}_{U}^{pool}$
\State Randomly chose the layer index $k$ of the $f_{\theta}$
\For {$i=0,1,2,\ldots$}\textcolor{gray}{\Comment{for the unlabeled instances}}
\State $x_{i}\in\mathcal{X}_{U}^{pool}$, $x'_i\in\mathcal{X'}_{U}^{pool}$
\State Get $h^{k}(x_{i}),h^{k}(x'_{i})$ of $x_{i}, x'_{i}$ as Eq.\eqref{q;tilde_x}
\For {$q = 0,1,2,\ldots$} \textcolor{gray}{\Comment{training of $\pi_{\phi}$}}
\State $\tau^{q}_{i} = \pi_{\phi}([h^{k}(x_{i});h^{k}(x'_i)])$
\State Calculate $\frac{\partial}{\partial \phi}L_{\pi}$ as Eq.\eqref{q;L_pi}
\State $\phi \leftarrow \phi - \eta_{\pi} \frac{\partial}{\partial \phi} L_{\pi}$
\If {$L_{\pi}$ is minimal}
\State {$\tau^{*}_{i}=\tau^{q}_{i}$}
\EndIf
\EndFor
\EndFor
\State Select and query the dataset, $\mathcal{X}_{S}$, as Eq.\eqref{eq:xs}
\State $\mathcal{X}_{L}^{j+1}=\mathcal{X}_{L}^{j} \cup \mathcal{X}_{S}$
\For {$t=0,1,2,\ldots$} \textcolor{gray}{\Comment{training of $f_{\theta}$}}
\State $\lambda_{i} \sim \text{Beta}(\tau_{i}^{*},\tau_{i}^{*})$ for $(x_i,x'_{i}) \in \mathcal{X}_{S}$
\State Get virtual dataset, $\mathcal{X}_{M}$, as Eq.\eqref{eq:xm}
\State Calculate $L_f$ as Eq.\eqref{eq:Ll} $\sim$ \eqref{eq:Lm}
\State $\theta \leftarrow \theta - \eta_{f} \frac{\partial}{\partial \theta} L_f$
\EndFor
\EndFor
\end{algorithmic}
\end{algorithm}

\subsection{Training Set Expansion through Acquisition}
We assume that we start the $j^{th}$ active learning iteration with already acquired labeled dataset $\mathcal{X}_{L}^{j}$. With the allowed budget per acquisition as $k$, we acquire top-$\frac{k}{2}$ pairs, i.e. $\mathcal{X}_{S}$ among the subsets, $\mathcal{X}'_{S}\subset\mathcal{X}_{U}\times\mathcal{X}_{U}$, with $|\mathcal{X}'_{S}|=\frac{k}{2}$.
\begin{equation}
    \mathcal{X}_{S}=\operatorname*{argmax}_{\mathcal{X}'_{S}\subset\mathcal{X}_{U}\times\mathcal{X}_{U}} \sum_{(x_i,x'_i)\in\mathcal{X}'_{S}}f_{acq}((x_i,x'_i);f_{\theta})\label{eq:xs}
\end{equation}
At this moment, \textit{oracle} annotates the true labels on $\mathcal{X}_{S}$.
Also, we have a virtual instance dataset, $\mathcal{X}_{M}$, generated by \textit{InfoMixup} with the optimal mixing policy, $\tau^*$:
\begin{equation}
\mathcal{X}_{M}=\bigcup_{(x_i,x'_i)\in\mathcal{X}_{S}}\{\lambda_{i}f_{\theta}^{0:L}(x_i)+(1-\lambda_{i})f_{\theta}^{0:k}(x'_i)\},\label{eq:xm}
\end{equation}
where $\lambda_i\sim \text{Beta}(\tau_i^*,\tau_i^*)$. Here, $\tau^*$ is dynamically inferred by the neural network of $\pi_{\phi}$ per each pair.

Up to this phase, our training dataset becomes $\mathcal{X}_{L}^{j+1}=\mathcal{X}_{L}^{j} \cup \mathcal{X}_{S}$ and $\mathcal{X}_{M}$.
Our proposed algorithm, described in Algorithm \ref{algorithm_InfoMixupAcq}, utilizes $\mathcal{X}_{M}$ for this active learning iteration only, with various $\lambda_i$s sampled at each training epoch. The classifier network's parameter, $\theta$, is learned via the gradient of the cross-entropy loss,
\begin{align}
L_f &= CE(f_{\theta}(x_{i}),y_{i})_{x_i\in\mathcal{X}_{L}^{j+1}}\label{eq:Ll}\\
&+CE(f_{\theta}^{k:L}(x_i),y_i)_{x_i\in\mathcal{X}_{M}},\label{eq:Lm}
\end{align}
where $y_i$ is the corresponding ground-truth label annotated from the \textit{oracle} for Eq.\eqref{eq:Ll}; or the mixed label according to the mixing policy for Eq.\eqref{eq:Lm}.

\subsection{\text{LADA} with Various Augmentation-Acquisition}
Since we propose the integrated framework of acquisition function and data augmentation to look ahead the informativeness of the data, we can use various acquisition functions and data augmentations in our \text{LADA} framework.
For example, we may substitute the \textit{Max Entropy}, which is the feedback of the acquisition function to the data augmentation in \textit{InfoMixup}, with another simple feedback, Var Ratio. Also, if we apply the VAAL acquisition function, \text{LADA} with VAAL trains the generator network, $\pi_{\phi}$, to maximize the discriminator's indication on the unlabeled dataset, $\text{P}(x\in\mathcal{X}_{U};\varphi^{VAAL})$, for the generated instances.

Similarly, we may substitute the data augmentation of \textit{InfoMixup} with Spatially Transform Networks (STN) \cite{STN}, a.k.a. \textit{InfoSTN}. STN may be trained with a subset of unlabeled data as input to maximize their predictive entropy when propagated to the current classifier network. The score to pick the most informative data is formulated as $f_{acq}^{STN}(x_i;f_{\theta}) = \mathbb{H}[\hat{y}^{STN}_i|x^{STN}_{i}]$, where $x^{STN}_{i}$ is the spatially transformed output of the data $x_{i}$, and $\hat{y}^{STN}_i$ is the corresponding prediction by the current classifier network. We provide more details in Appendix C.

\section{Experiments}\label{Section_Experiments}
\subsection{Baselines and Datasets}
This section denotes the proposed framework as \text{LADA}, and we specify the instantiated data augmentation and acquisition by its subscript, i.e. the proposed InfoMixup as $\mathrm{LADA_{EntMix}}$ which adopts \textit{Max Entropy} as data acquisition and \textit{Mixup} as data augmentation to select and generate \textit{informative} samples.
If we change the entropy measure to the Var Ratio or the discriminator logits of VAAL, it results in the subscript of \text{VarMix} or \text{VaalMix}, respectively. Also, if we change the augmentation policy to the STN network, the subscript becomes \text{EntSTN}.

We compare our models to 1) Coreset \cite{coreset}; 2) BADGE \cite{badge}; and 3) VAAL \cite{vaal} as the baselines for active learning. We also include some data augmented active learning: 1) BGADL, 2) \textit{Manifold Mixup}; and 3) \textit{AdaMixup}. Here, BGADL is an integrated data augmentation and acquisition method, but it should be noted that BGADL has no learning mechanism in the augmentation from the feedback of acquisition.
We also add ablated baselines to see the effect of learning $\tau$, so we introduce the fixed $\tau$ case as $\mathrm{LADA^{fixed}}$.
The classifier network, $f_{\theta}$, adopts Resnet-18 \cite{resnet}, and the policy generator network, $\pi_{\phi}$, consists of a much smaller neural network. Appendix A.2 provides more details on the networks and their training.

We experiment the above-mentioned models with three benchmark datasets: FashionMNIST (Fashion) \cite{fashion}, SVHN \cite{svhn}, and CIFAR-10 \cite{cifar10}.
Throughout our experiments, we repeat the experiments for five times to validate the statistical significance, and the maximum acquisition iteration is limited to 100. More details about the treatment on each dataset are in Appendix A.1.

We evaluate the models under the pool-based active learning scenario. We assume that the model has 20 training instances, which are randomly chosen and balanced. As the active learning iteration progresses, we acquire 10 additional training instances at each iteration, and we use the same amount of \textit{oracle} queries for all models, which results in selecting top-5 pairs when adopting \textit{Mixup} as data augmentation in the \text{LADA} framework.

\renewcommand{\arraystretch}{1.11}
\begin{table*}[h]
  \begin{varwidth}[b]{0.75\linewidth}
    \centering
    \begin{tabular}{c|c||ccc||cc}
\hline
\multicolumn{2}{c||}{Method}                                        & Fashion        & SVHN           & CIFAR-10        & Time  & Param.\\ \hline
\multirow{6}{*}{\rotatebox{90}{Baselines}}      & Random            & 80.96$\pm$0.62 & 73.92$\pm$2.80 & 35.27$\pm$1.36  & 1     & -\\
                                                & BALD              & 80.99$\pm$0.59 & 75.66$\pm$2.07 & 34.71$\pm$2.28  & 1.36  & -\\
                                                & Coreset           & 78.47$\pm$0.30 & 68.57$\pm$3.13 & 28.25$\pm$0.89  & 1.54  & -\\
                                                & BADGE             & 80.94$\pm$0.98 & 70.89$\pm$1.91 & 28.60$\pm$1.17  & 1.31  & -\\
                                                & BGADL             & 78.42$\pm$1.05 & 63.50$\pm$1.56 & 35.08$\pm$2.20  & 4.69  & 13M\\ \hline
\multirow{7}{*}{\rotatebox{90}{Entropy-based}}  & \textit{Max Entropy}              & 80.93$\pm$1.85 & 72.57$\pm$0.76 & 34.97$\pm$0.71  & 1.01 & -\\
                                                & Ent w.\textit{ManifoldMixup}      & 82.31$\pm$0.38 & 72.69$\pm$1.29 & 35.88$\pm$0.85  & 1.03 & -\\
                                                & Ent w.\textit{AdaMixup}           & 81.30$\pm$0.83 & 73.00$\pm$0.39 & 35.67$\pm$1.75  & 1.03 & 5K\\
                                                & $\mathrm{LADA^{fixed}_{EntMix}}$         & 83.08$\pm$1.34 & 75.73$\pm$1.48 & 36.34$\pm$0.88  & 1.06 & -\\
                                                & $\mathrm{LADA_{EntMix}}$               & \textbf{83.67$\pm$0.29} & \textbf{76.55$\pm$0.31} & \textbf{37.04$\pm$1.34} & 1.32 & 77K\\ \cline{2-7} 
                                                & $\mathrm{LADA^{fixed}_{EntSTN}}$         & 82.37$\pm$0.58 & 72.08$\pm$1.67 & 35.55$\pm$1.34 & 1.02 & 5K\\
                                                & $\mathrm{LADA_{EntSTN}}$               & 81.83$\pm$0.55 & 73.80$\pm$0.81 & 36.18$\pm$0.69 & 1.20 & 5K\\ \hline
\multirow{3}{*}{\rotatebox{90}{VAAL}\rotatebox{90}{-based}}     & VAAL                              & \textbf{82.67$\pm$0.29} & 75.01$\pm$0.66 & 39.82$\pm$0.86 & 3.55 & 301K\\
                                                                & $\mathrm{LADA^{fixed}_{VaalMix}}$        & 82.63$\pm$0.29 & 76.83$\pm$1.05 & 44.42$\pm$2.12 & 3.56 & 301K\\
                                                                & $\mathrm{LADA_{VaalMix}}$              & 82.60$\pm$0.49 & \textbf{77.92$\pm$0.51} & \textbf{44.56$\pm$1.40} & 3.60 & 378K\\ \hline
\multirow{3}{*}{\rotatebox{90}{VarRatio}\rotatebox{90}{-based}} & Var Ratio                        & 81.05$\pm$0.18 & 74.07$\pm$1.87 & 34.99$\pm$0.73 & 1.01 & -\\
                                                                & $\mathrm{LADA^{fixed}_{VarMix}}$         & 83.11$\pm$0.66 & 76.01$\pm$2.64 & 35.98$\pm$1.68 & 1.06 & -\\
                                                                & $\mathrm{LADA_{VarMix}}$               & \textbf{84.47$\pm$0.89} & \textbf{76.09$\pm$0.94} & \textbf{36.84$\pm$0.51} & 1.33 & 77K\\ \hline
\end{tabular}
    \caption{Comparison of test accuracy, the run-time of 1 iteration of acquisition (Time), and the number of parameters (Param.). The best performance in each category is indicated as boldface. The run-time is calculated as the ratio to the Random acquisition. The number of parameters is only reported for the auxiliary network, and - indicates no auxiliary network in the method.}
  \end{varwidth}
  \hfill
  \begin{minipage}[b]{0.23\linewidth}
    \centering
    \begin{subfigure}{1.0\linewidth}
        \centering 
        \includegraphics[width=40mm,height=25.3mm]{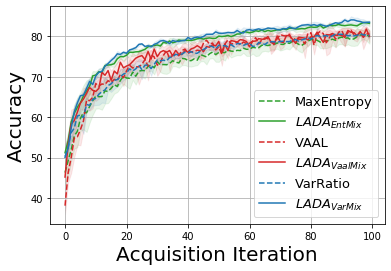}
        \vspace{-1mm}
        \caption{Fashion}
        \label{fig_AccFashion}
    \end{subfigure}
    \begin{subfigure}{1.0\linewidth}
        \centering 
        \includegraphics[width=40mm,height=25.3mm]{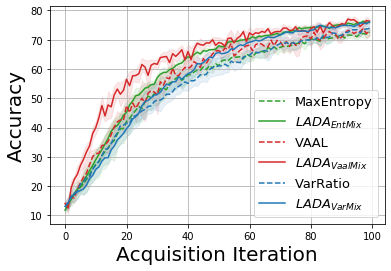}
        \vspace{-1mm}
        \caption{SVHN}
        \label{fig_AccSVHN}
    \end{subfigure}
    \begin{subfigure}{1.0\linewidth}
        \centering 
        \includegraphics[width=40mm,height=25.3mm]{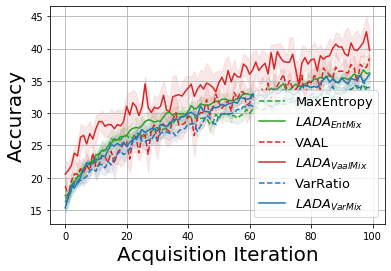}
        \vspace{-1mm}
        \caption{CIFAR-10}
        \label{fig_AccCifar10}
    \end{subfigure}
        \captionof{figure}{Test accuracy over the acquisition iterations}
        \label{fig_TestAcc_Total}
  \end{minipage}
\end{table*}

\subsection{Quantitative Performance Evaluations}
Table \ref{fig_TestAcc_Total} shows the average test accuracy, and the accuracy of each replication represents the best accuracy over the acquisition iterations. Since we introduce the generalizable framework, Table \ref{fig_TestAcc_Total} separates the performances by the instantiated acquisition functions. The group of baselines does not have any learning mechanism on the acquisition metric, and this group has the worst performances. We suggest three acquisition functions to be adopted by our \text{LADA} framework, which are 1) the predictive entropy by the classifier, 2) the discriminator logits in VAAL, and 3) the classifier variation ratio. Given that VAAL uses a discriminator and a generator, the VAAL-based model has more parameters to optimize in terms of complexity, which provides an advantage in a complex dataset, such as CIFAR-10.

When we examine the general performance gains across datasets, we find the best performers as $\mathrm{LADA_{VarMix}}$ in Fashion; and $\mathrm{LADA_{VaalMix}}$ in SVHN and CIFAR-10. In terms of the data augmentation, \textit{Mixup}-based augmentation outperforms STN augmentation.
As the dataset becomes complex, the greater gain of performance is achieved by $\mathrm{LADA_{EntSTN}}$ in SVHN or CIFAR-10, compared to Fashion.
In all combinations of baselines and datasets, the integrations of augmentation and acquisition, a.k.a. \text{LADA} variations, show the best performance in most cases. In terms of the ablation study, the learning of the data augmentation policy, $\tau$, is meaningful because the 10 learning case of \text{LADA} is better than the fixed case in 12 variations of \text{LADA}. Figure \ref{fig_TestAcc_Total} shows the convergence speed to the best test accuracy by each model. As the dataset becomes complex, the performance gain by \text{LADA} becomes apparent.

Additionally, we compare the integrated framework to the pipelined approach. \textit{Max Entropy} does not have an augmentation part, so it becomes the simplest model. Then, Ent w.\textit{Manifold Mixup} adds the \textit{Manifold Mixup} augmentation, but it does not have a learning process on the mixing policy. Finally, Ent w.\textit{AdaMixup} has a learning process on the mixing policy, but the learning is separated from the acquisition. These pipelined approaches show lower performances than the integration cases of \text{LADA}.

\begin{figure}[h]
\centering
\vspace{-2mm}
\begin{subfigure}{.49\linewidth}
    \centering 
    \includegraphics[width=.95\linewidth]{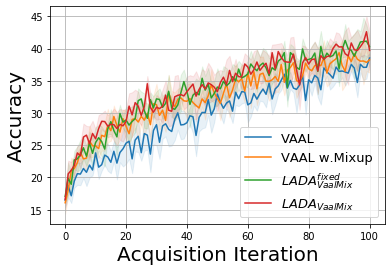}
    \caption{$\mathrm{LADA_{VaalMix}}$}
    \label{fig_AcqModule}
\end{subfigure}
\begin{subfigure}{.49\linewidth}
    \centering 
    \includegraphics[width=.95\linewidth]{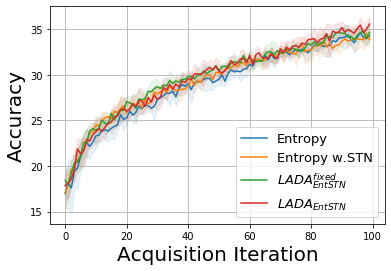}
    \caption{$\mathrm{LADA_{EntSTN}}$}
    \label{fig_DaModule}
\end{subfigure}
\vspace{-2mm}
\caption{Ablation study of \text{LADA} with various acquisition and augmentation on CIFAR-10 dataset}
\label{fig_Module}
\end{figure}

Finally, as \text{LADA} is a generalizable framework to work with the various acquisition and augmentation functions, Figure \ref{fig_AcqModule} and Figure \ref{fig_DaModule} show the ablation study on the instantiated \text{LADA} with the VAAL acquisition function and the $STN$ augmentation function, respectively. The figures confirm the effects of both integration and learnable augmentation policy with feedback from the acquisition.

\vspace{-2mm}

\subsection{Qualitative Analysis on Acquired Data Instances}
Besides the quantitative comparison, we need reasoning on the behavior of \text{LADA}. Therefore, we selected $\mathrm{LADA_{EntMix}}$ to contrast to the pipelined approach. We investigate on 1) achieving the informative data instances by acquisition, 2) the validity of the optimal $\tau^{*}$ in the augmentation learned from the policy generator network $\pi_{\phi}$, and 3) examining the coverage of the explored space.

\begin{figure}[t]
\centering
\begin{subfigure}{.49\linewidth}
    \centering 
    \includegraphics[width=.95\linewidth]{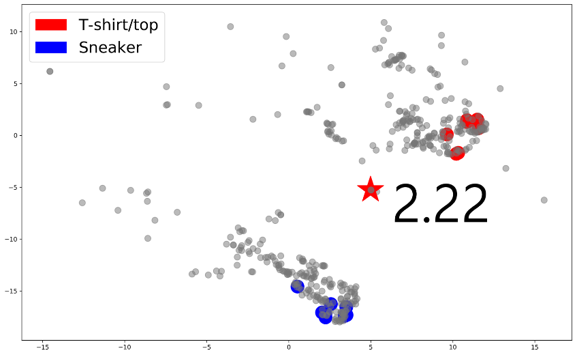}
    \vspace{-1mm}
    \caption{\textit{Max Entropy} at early}
    \label{fig_AcqOnlyEarly}
\end{subfigure}
\begin{subfigure}{.49\linewidth}
    \centering 
    \includegraphics[width=.95\linewidth]{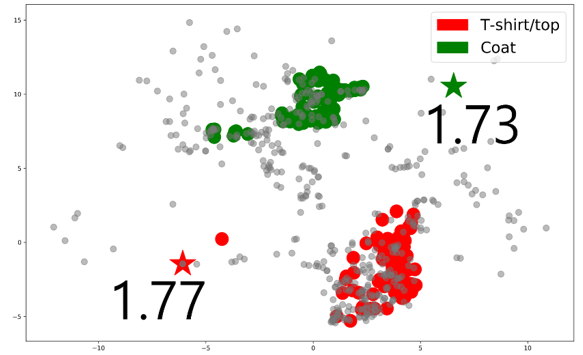}
    \vspace{-1mm}
    \caption{\textit{Max Entropy} at late}
    \label{fig_AcqOnlyLater}
\end{subfigure}

\begin{subfigure}{.49\linewidth}
    \centering
    \includegraphics[width=.95\linewidth]{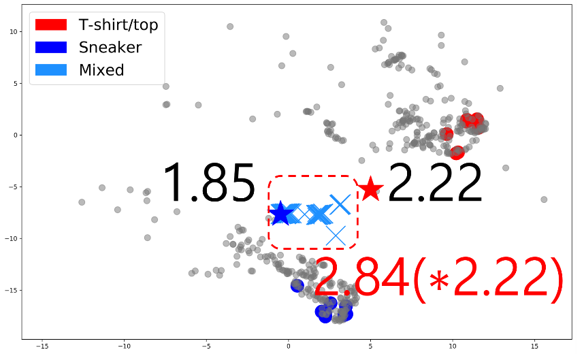}
    \vspace{-1mm}
    \caption{$\mathrm{LADA_{EntMix}}$ at early}
    \label{fig_InfoEarly}
\end{subfigure}
\begin{subfigure}{.49\linewidth}
    \centering
    \includegraphics[width=.95\linewidth]{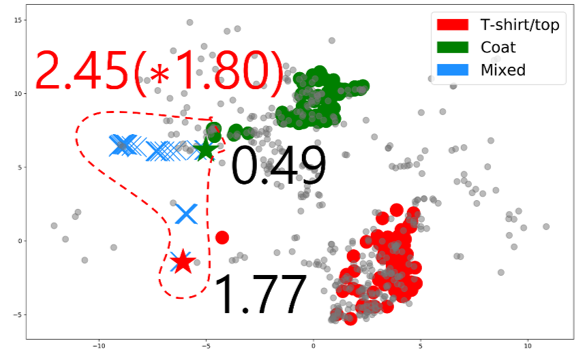}
    \vspace{-1mm}
    \caption{$\mathrm{LADA_{EntMix}}$ at late}
    \label{fig_InfoLater}
\end{subfigure}
\vspace{-2mm}
\caption{tSNE \cite{tsne} plot of acquired instance ($\star$) and augmented instance ($\times$), with entropy values. The numbers written in \textit{black} indicate the predictive entropy of unlabeled data instances that were selected from the unlabeled pool. The numbers written in \textit{red} indicate the maximum (*average) value of predictive entropy of the virtual data instances that were generated from \textit{InfoMixup}. The acquisition iterations of early and late are 7 and 76.}
\label{fig_Acq_Info}
\end{figure}

To check the informativeness of data instances, Figure \ref{fig_Acq_Info} shows the different acquisition process between \textit{Max Entropy} and $\mathrm{LADA_{EntMix}}$. \textit{Max Entropy} selects a data instance with the highest predictive entropy value. Compared to \textit{Max Entropy}, $\mathrm{LADA_{EntMix}}$ selects a pair of two data instances with the highest value of the aggregated predictive entropy, which is the summation of the predictive entropy from two data instances and one $\textit{InfoMixup}$ instance.
By mixing two unlabeled data instances with the corresponding optimal mixing policy $\tau^{*}$, the virtual data instance, generated along with the vicinal space, results in a high entropy value, which can be higher than the selected instance by \textit{Max Entropy}. The virtual data instance helps the current classifier model to clarify the decision boundary between two classes along the interpolation line of two mixed real instances.

\begin{figure}[h]
\captionsetup[subfigure]{justification=centering}
\centering
\begin{subfigure}{.49\linewidth}
    \centering 
    \includegraphics[scale=0.23]{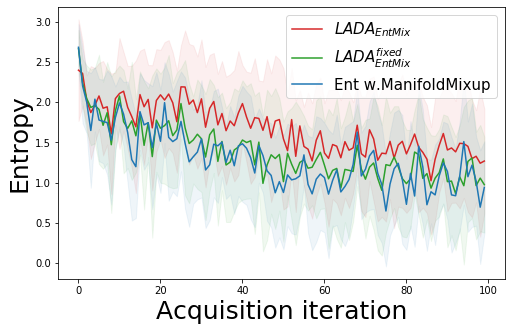}
    \caption{Mean value of entropy \\ of the virtual data}
    \label{fig_Entropy}
\end{subfigure}
\begin{subfigure}{.49\linewidth}
    \centering 
    \includegraphics[scale=0.23]{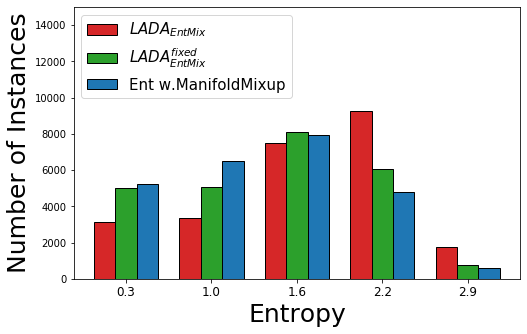}
    \caption{Number of virtual data \\ with entropy values}
    \label{fig_Histogram}
\end{subfigure}
\vspace{-2mm}
\caption{Entropy values of the virtual data generated from $\mathrm{LADA_{EntMix}}$, $\mathrm{LADA^{fixed}_{EntMix}}$, and Ent w.\textit{Manifold Mixup}}
\label{fig_Informativeness}
\end{figure}

To confirm the validity of the optimal $\tau^{*}$, we compare three cases of 1) the inferred $\tau$ ($\mathrm{LADA_{EntMix}}$); 2) the fixed $\tau$ ($\mathrm{LADA^{fixed}_{EntMix}}$); and 3) the pipelined model's $\tau$ (Ent w.\textit{Manifold Mixup}).
Figure \ref{fig_Entropy} shows the entropy of the virtual data instances over the acquisition process. As expected, the optimal $\tau^{*}$ learned in $\mathrm{LADA_{EntMix}}$ produces the highest entropy over the acquisition process, but it should be noted that the differentiation becomes significant after some acquisition iterations, which comes from the requirement of training the classifier.
Figure \ref{fig_Histogram} shows the distribution of entropy of virtual instances, with the median value of each interval as $x$-axis.
This also shows that the optimal $\tau^{*}$ has the highest density beyond the interval of the median 2.2.

\begin{figure}[h]
\centering
\begin{subfigure}{.49\linewidth}
    \centering 
    \includegraphics[width=.95\linewidth]{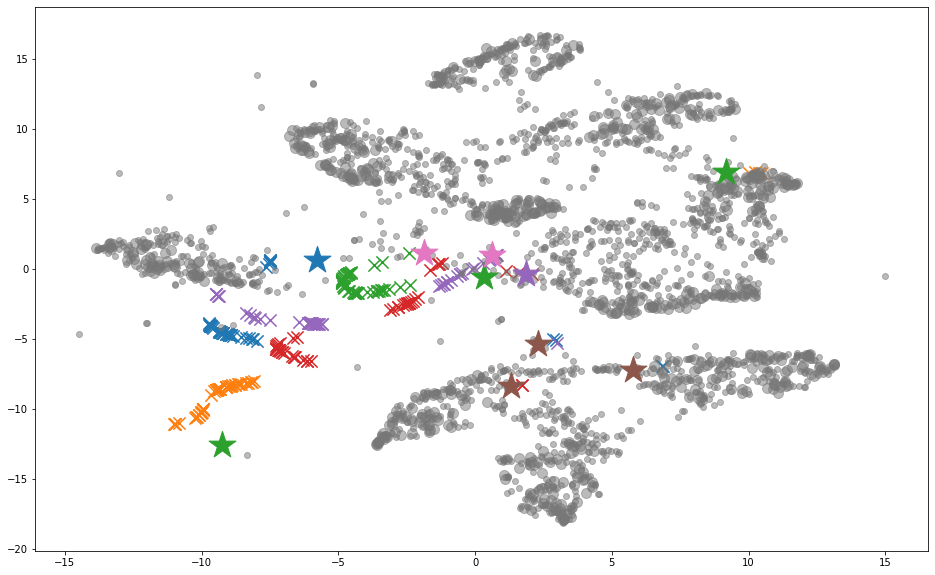}
    \caption{$\mathrm{LADA_{EntMix}}$}
    \label{fig_distance_InfoMixup}
\end{subfigure}
\begin{subfigure}{.49\linewidth}
    \centering 
    \includegraphics[width=.95\linewidth]{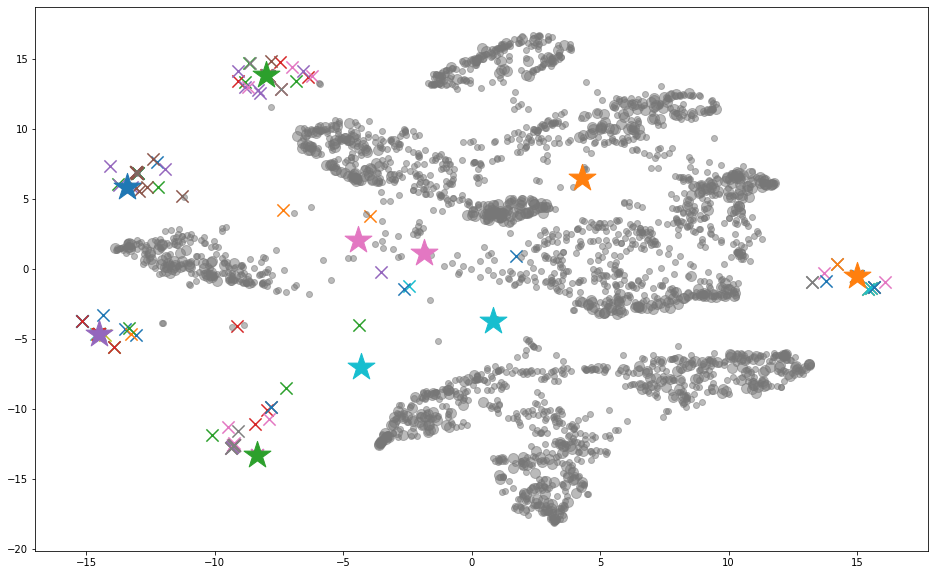}
    \caption{Ent w.\textit{AdaMixup}}
    \label{fig_distance_AdaMixup}
\end{subfigure}
\vspace{-2mm}
\caption{tSNE plot of acquired data instances ($\star$) and generated virtual data instances ($\times$). The labels are categorized by colors.}
\label{fig_Distance}
\end{figure}

To examine the coverage of the explored latent space, Figure \ref{fig_Distance} illustrates the latent space of the acquired data instances and the augmented data instances. Ent w.\textit{AdaMixup} has a potential capability of interpolating distantly paired data instances, but its learned $\tau$ limits a sample of $\lambda$ to be placed near either one of the paired instances because of the aversion on the manifold intrusion. Therefore, in the experiments, Ent w.\textit{AdaMixup} ends up exploring the space near the acquired instances.
The generated virtual data instances by $\mathrm{LADA_{EntMix}}$ show further exploration than Ent w.\textit{AdaMixup}. The latent space makes the linear interpolation of $\mathrm{LADA_{EntMix}}$ to be curved by the manifold, but it keeps the interpolation line of the curved manifold. The extent of the interpolation is broader than \textit{AdaMixup} because the optimal $\tau^{*}$ is guided by the entropy maximization, which is adversarial in a sense. This adversarial approach is different from the aversion of the manifold intrusion because the latter is more conservative to the currently learned parameter.

\vspace{-3mm}

\section{Conclusions}
In the real world where gathering a large-scale labeled dataset is difficult because of the constrained human or computational resources, learning the deep neural network requires an effective utilization of the limited resources. This limitation motivated the integration of data augmentation and active learning.
This paper proposes a generalized framework for such integration, named as \text{LADA}, which adaptively selects the informative data instances by looking ahead the acquisition score of both 1) the unlabeled data instances and 2) the virtual data instances to be generated by data augmentation, in advance of the acquisition process. To enhance the effect of the data augmentation, \text{LADA} learns the augmentation policy to maximize the acquisition score.
With repeated experiments on the various datasets and the comparison models, \text{LADA} shows a considerable performance by selecting and augmenting informative data instances. The qualitative analysis shows the different behavior of \text{LADA} that finds the vicinal space of high acquisition score by learning the optimal policy.

\section{Ethics Statement}
In the real world, the limited amount of labeled dataset makes it hard to train the deep neural networks and high cost of the annotation cost becomes problematic. This leads to the decision on what to select and annotate first, which calls upon the active learning. Besides the active learning, effectively enlarging the limited amount of labeled dataset is also considerable.
With this motivations, we propose a framework that can adopt various types of acquisitions and augmentations that exist in machine learning field. By looking ahead the effect of data augmentation in the process of acquisition, we can select data instances that are informative if selected and labeled but also augmented. Moreover, by learning the augmentation policy in advance of the actual acquisition process, we enhance the informativeness of the generated virtual data instances.
We believe that the proposed LADA framework can improve the performance of deep learning models, especially when the annotation by human experts is expensive.

\bibliographystyle{aaai21}
\bibliography{LADAbib}

\begin{thebibliography}{28}
\providecommand{\natexlab}[1]{#1}
\providecommand{\url}[1]{\texttt{#1}}
\providecommand{\urlprefix}{URL }
\expandafter\ifx\csname urlstyle\endcsname\relax
  \providecommand{\doi}[1]{doi:\discretionary{}{}{}#1}\else
  \providecommand{\doi}{doi:\discretionary{}{}{}\begingroup
  \urlstyle{rm}\Url}\fi

\bibitem[{Ash et~al.(2020)Ash, Zhang, Krishnamurthy, Langford, and
  Agarwal}]{badge}
Ash, J.~T.; Zhang, C.; Krishnamurthy, A.; Langford, J.; and Agarwal, A. 2020.
\newblock Deep Batch Active Learning by Diverse, Uncertain Gradient Lower
  Bounds.
\newblock In \emph{ICLR}.

\bibitem[{Chapelle et~al.(2001)Chapelle, Weston, Bottou, and Vapnik}]{VRM}
Chapelle, O.; Weston, J.; Bottou, L.; and Vapnik, V. 2001.
\newblock Vicinal risk minimization.
\newblock In \emph{Advances in neural information processing systems},
  416--422.

\bibitem[{Cohn, Ghahramani, and Jordan(1996)}]{cohn1996active}
Cohn, D.~A.; Ghahramani, Z.; and Jordan, M.~I. 1996.
\newblock Active learning with statistical models.
\newblock \emph{Journal of artificial intelligence research} 4: 129--145.

\bibitem[{Cubuk et~al.(2019)Cubuk, Zoph, Mane, Vasudevan, and Le}]{autoaugment}
Cubuk, E.~D.; Zoph, B.; Mane, D.; Vasudevan, V.; and Le, Q.~V. 2019.
\newblock Autoaugment: Learning augmentation strategies from data.
\newblock In \emph{Proceedings of the IEEE conference on computer vision and
  pattern recognition}, 113--123.

\bibitem[{Freeman(1965)}]{VarRatio}
Freeman, L. 1965.
\newblock \emph{Elementary applied statistics: for students in behavioral
  science}.
\newblock Wiley.
\newblock \urlprefix\url{https://books.google.co.kr/books?id=r4VRAAAAMAAJ}.

\bibitem[{Goodfellow et~al.(2014)Goodfellow, Pouget-Abadie, Mirza, Xu,
  Warde-Farley, Ozair, Courville, and Bengio}]{GAN}
Goodfellow, I.~J.; Pouget-Abadie, J.; Mirza, M.; Xu, B.; Warde-Farley, D.;
  Ozair, S.; Courville, A.; and Bengio, Y. 2014.
\newblock Generative Adversarial Nets.
\newblock In \emph{Proceedings of the 27th International Conference on Neural
  Information Processing Systems - Volume 2}, NIPS’14, 2672–2680.
  Cambridge, MA, USA: MIT Press.

\bibitem[{Guo, Mao, and Zhang(2019)}]{AdaMixup}
Guo, H.; Mao, Y.; and Zhang, R. 2019.
\newblock Mixup as locally linear out-of-manifold regularization.
\newblock In \emph{Proceedings of the AAAI Conference on Artificial
  Intelligence}, volume~33, 3714--3722.

\bibitem[{Hastings(1970)}]{hastings1970monte}
Hastings, W.~K. 1970.
\newblock Monte Carlo sampling methods using Markov chains and their
  applications .

\bibitem[{He et~al.(2016)He, Zhang, Ren, and Sun}]{resnet}
He, K.; Zhang, X.; Ren, S.; and Sun, J. 2016.
\newblock Deep residual learning for image recognition.
\newblock In \emph{Proceedings of the IEEE conference on computer vision and
  pattern recognition}, 770--778.

\bibitem[{Houlsby et~al.(2011)Houlsby, Huszar, Ghahramani, and Lengyel}]{BALD}
Houlsby, N.; Huszar, F.; Ghahramani, Z.; and Lengyel, M. 2011.
\newblock Bayesian Active Learning for Classification and Preference Learning.
\newblock \emph{CoRR} abs/1112.5745.

\bibitem[{Jaderberg et~al.(2015)Jaderberg, Simonyan, Zisserman et~al.}]{STN}
Jaderberg, M.; Simonyan, K.; Zisserman, A.; et~al. 2015.
\newblock Spatial transformer networks.
\newblock In \emph{Advances in neural information processing systems},
  2017--2025.

\bibitem[{Jankowiak and Karaletsos(2019)}]{JankowiakK19}
Jankowiak, M.; and Karaletsos, T. 2019.
\newblock Pathwise Derivatives for Multivariate Distributions.
\newblock In Chaudhuri, K.; and Sugiyama, M., eds., \emph{The 22nd
  International Conference on Artificial Intelligence and Statistics, {AISTATS}
  2019, 16-18 April 2019, Naha, Okinawa, Japan}, volume~89 of \emph{Proceedings
  of Machine Learning Research}, 333--342. {PMLR}.
\newblock \urlprefix\url{http://proceedings.mlr.press/v89/jankowiak19a.html}.

\bibitem[{Jankowiak and Obermeyer(2018)}]{jankowiak2018pathwise}
Jankowiak, M.; and Obermeyer, F. 2018.
\newblock Pathwise Derivatives Beyond the Reparameterization Trick.
\newblock In Dy, J.~G.; and Krause, A., eds., \emph{Proceedings of the 35th
  International Conference on Machine Learning, {ICML} 2018,
  Stockholmsm{\"{a}}ssan, Stockholm, Sweden, July 10-15, 2018}, volume~80 of
  \emph{Proceedings of Machine Learning Research}, 2240--2249. {PMLR}.
\newblock \urlprefix\url{http://proceedings.mlr.press/v80/jankowiak18a.html}.

\bibitem[{Kingma and Welling(2014)}]{VAE}
Kingma, D.~P.; and Welling, M. 2014.
\newblock Auto-Encoding Variational Bayes.
\newblock In Bengio, Y.; and LeCun, Y., eds., \emph{2nd International
  Conference on Learning Representations, {ICLR} 2014, Banff, AB, Canada, April
  14-16, 2014, Conference Track Proceedings}.
\newblock \urlprefix\url{http://arxiv.org/abs/1312.6114}.

\bibitem[{Krizhevsky, Hinton et~al.(2009)}]{cifar10}
Krizhevsky, A.; Hinton, G.; et~al. 2009.
\newblock Learning multiple layers of features from tiny images .

\bibitem[{Liu and Ferrari(2017)}]{poseestimation}
Liu, B.; and Ferrari, V. 2017.
\newblock Active learning for human pose estimation.
\newblock In \emph{Proceedings of the IEEE International Conference on Computer
  Vision}, 4363--4372.

\bibitem[{Maaten and Hinton(2008)}]{tsne}
Maaten, L. v.~d.; and Hinton, G. 2008.
\newblock Visualizing data using t-SNE.
\newblock \emph{Journal of machine learning research} 9(Nov): 2579--2605.

\bibitem[{Netzer et~al.(2011)Netzer, Wang, Coates, Bissacco, Wu, and Ng}]{svhn}
Netzer, Y.; Wang, T.; Coates, A.; Bissacco, A.; Wu, B.; and Ng, A.~Y. 2011.
\newblock Reading digits in natural images with unsupervised feature learning .

\bibitem[{Perez and Wang(2017)}]{perez2017effectiveness}
Perez, L.; and Wang, J. 2017.
\newblock The effectiveness of data augmentation in image classification using
  deep learning.
\newblock \emph{arXiv preprint arXiv:1712.04621} .

\bibitem[{Sener and Savarese(2018)}]{coreset}
Sener, O.; and Savarese, S. 2018.
\newblock Active Learning for Convolutional Neural Networks: A Core-Set
  Approach.
\newblock In \emph{International Conference on Learning Representations}.

\bibitem[{Settles(2009)}]{settles2009active}
Settles, B. 2009.
\newblock Active learning literature survey.
\newblock Technical report, University of Wisconsin-Madison Department of
  Computer Sciences.

\bibitem[{Shannon(1948)}]{maxentropy}
Shannon, C.~E. 1948.
\newblock A mathematical theory of communication.
\newblock \emph{Bell Syst. Tech. J.} 27(3): 379--423.

\bibitem[{Sinha, Ebrahimi, and Darrell(2019)}]{vaal}
Sinha, S.; Ebrahimi, S.; and Darrell, T. 2019.
\newblock Variational adversarial active learning.
\newblock In \emph{Proceedings of the IEEE International Conference on Computer
  Vision}, 5972--5981.

\bibitem[{Tong(2001)}]{tong2001active}
Tong, S. 2001.
\newblock \emph{Active learning: theory and applications}, volume~1.
\newblock Stanford University USA.

\bibitem[{Tran et~al.(2019)Tran, Do, Reid, and Carneiro}]{BGADL}
Tran, T.; Do, T.; Reid, I.~D.; and Carneiro, G. 2019.
\newblock Bayesian Generative Active Deep Learning.
\newblock In Chaudhuri, K.; and Salakhutdinov, R., eds., \emph{Proceedings of
  the 36th International Conference on Machine Learning, {ICML} 2019, 9-15 June
  2019, Long Beach, California, {USA}}, volume~97 of \emph{Proceedings of
  Machine Learning Research}, 6295--6304. {PMLR}.
\newblock \urlprefix\url{http://proceedings.mlr.press/v97/tran19a.html}.

\bibitem[{Verma et~al.(2018)Verma, Lamb, Beckham, Najafi, Courville,
  Mitliagkas, and Bengio}]{ManifoldMixup}
Verma, V.; Lamb, A.; Beckham, C.; Najafi, A.; Courville, A.; Mitliagkas, I.;
  and Bengio, Y. 2018.
\newblock Manifold mixup: Learning better representations by interpolating
  hidden states .

\bibitem[{Xiao, Rasul, and Vollgraf(2017)}]{fashion}
Xiao, H.; Rasul, K.; and Vollgraf, R. 2017.
\newblock Fashion-mnist: a novel image dataset for benchmarking machine
  learning algorithms.
\newblock \emph{arXiv preprint arXiv:1708.07747} .

\bibitem[{Zhang et~al.(2017)Zhang, Cisse, Dauphin, and Lopez-Paz}]{Mixup}
Zhang, H.; Cisse, M.; Dauphin, Y.~N.; and Lopez-Paz, D. 2017.
\newblock mixup: Beyond empirical risk minimization.
\newblock \emph{arXiv preprint arXiv:1710.09412} .

\end{thebibliography}

\end{document}